\documentclass[a4paper]{IEEEtran}
\usepackage{amsfonts,amssymb}
\usepackage{graphicx,epstopdf}
\usepackage{tikz}
\usetikzlibrary{automata}
\usepackage{color}
\usepackage{threeparttable}
\usepackage{amsmath}
\usepackage{amsfonts,amssymb}
\usepackage{slashbox}
\usepackage{epstopdf}
\usepackage[switch]{lineno}
\usepackage{algorithm}
\usepackage[noend]{algpseudocode}
\allowdisplaybreaks[2]

\usepackage[switch]{lineno}
\usepackage{etoolbox}
\patchcmd{\subequations}{}%
{}{}{}

\makeatletter
\def\BState{\State\hskip-\ALG@thistlm}
\makeatother

\IEEEoverridecommandlockouts   

\usepackage{epsfig}

\begin{document}
%



\title{Bicycle Detection Based On Multi-feature and Multi-frame Fusion in low-resolution traffic videos}

\author{Yicheng~Zhang,~\IEEEmembership{Student~Member,~IEEE,}
	 and~Qiang~Ling,~\IEEEmembership{Senior~Member,~IEEE}
}

\maketitle


%

\begin{abstract}
As a major type of transportation equipments, bicycles, including electrical bicycles, are distributed almost everywhere in China.
The accidents caused by bicycles have become a serious threat to the public safety.
So bicycle detection is one major task of traffic video surveillance systems in China.
In this paper, a method based on multi-feature and multi-frame fusion is presented for bicycle detection in low-resolution traffic videos.
It first extracts some geometric features of objects from each frame image, then concatenate multiple features into a feature vector and use linear support vector machine (SVM) to learn a classifier, or put these features into a cascade classifier, to yield a preliminary detection result regarding whether
an object is a bicycle.
It further fuses these preliminary detection results from multiple frames to
provide a more reliable detection decision, together with a confidence level of that decision.
Experimental results show that this method based on multi-feature and multi-frame fusion can identify
bicycles with high accuracy and low computational complexity.
It is, therefore, applicable for real-time traffic video surveillance systems.
\end{abstract}





\section{\textsc{Introduction}}
\label{sect:introduction}

In urban traffic systems, there are various participants, such as vehicles, pedestrians and bicycles\cite{Sheng2010}\cite{WangYubo2007}.
Bicycles are one of the major reasons of accidents in China.\cite{Ministry2011}.
In the last few years, we have seen much research on protecting vulnerable road users(VRUs), such as pedestrians, bicycles and other small vehicles,
which is a natural trend to enrich total driving safety\cite{Cho2010}.
Therefore it becomes a critical task to detect bicycles in urban intelligent traffic surveillance systems in order to reduce such accidents
\cite{Sheng2010}\cite{WangYubo2007}\cite{Sjt2010}.

There exist some bicycle detection methods, which can be classified into two types.
One type uses additional external sensors, such as laser sensors and infrared sensors \cite{WuYun2010}\cite{Niwat2011}.
The other type detects bicycles through image processing, which is the main focus of the present paper.
In \cite{Rogers1999}, one salient feature of bicycles, having two round wheels, is taken to detect a bicycle
through detecting two ellipses (the two wheels) in the image after Hough transformation.
In \cite{ChungChengChiu2007}, instead of wheels, the helmet of the bicycle rider is taken to
detect a bicycle.
The detection precision in \cite{Rogers1999} \cite{ChungChengChiu2007}, however, strongly depends on the video quality, and may be
poor for vague videos where it is quite difficult to detect a bicycle's two wheels or a bicycle rider's helmet.
In \cite{Cho2010}, a method is proposed to detect and track bicycle riders based on Histograms of Oriented Gradients (HOG).
The implementation of that method is limited because a HOG feature always requires large enough objects to ensure its accuracy
and effectiveness while the videos in real traffic video surveillance systems may not satisfy this requirement.
Moreover, it is time-consuming to extract HOG features due to their high complexity\cite{Dalal2006}.

Some other methods like using MSC-HOG method for detection \cite{Jung2012} or detecting tires of bicycles in videos \cite{Fujimoto2013} also can get good results, but they are either time consuming or high quality videos required. Some new methods, such the method based on HOG features with ROI in \cite{Moro2011}, try to use more advanced hardware device like GPU to finish the great amount of computation.

In summary, there are three major defects in the available bicycle detection methods based on image processing.
First, they require fine features for detection, which are hard to extract, particularly for traffic videos with low-resolution.
Second, the processing time under these methods is usually long and may not meet the requirement of the real-time detection.
Last, they make the bicycle detection decision by the information in a single frame, which may lead to misjudgment, especially in the case of strong noise or light changing.

Traffic videos, limited by their capture device and environmental conditions, have some prominent features, such as low resolution, complex background, various weather conditions and a variety of lighting levels.
Because of low-resolution, it is difficult to extract moving objects from the image precisely.
Subsequently, effective features that can be used in the object classification are hard to be obtained.

In order to resolve the above issues and to adapt to the application scenario of low-resolution traffic videos, we propose a bicycle detection method based on multi-feature and multi-frame fusion.
First, we extract geometric features and velocity features and then fuse them by using support vector machine (SVM) or cascade classifier.
As the objects are usually small in real video surveillance systems, we extract sparse geometric features, rather than dense features, to detect bicycles.
To enhance the precision of this feature descriptor, methods for feature fusion both on frame level and on image sequence level are proposed.
These multiple geometric features are concatenated into a feature vector and then the support vector machine (SVM) or
the cascade classifier methods are implemented to produce a preliminary detection decision
for the current single frame.
Moreover, we fuse the preliminary detection results from multiple frames by the majority rule,
which provides not only a more reliable detection result, but also the confidence level of that detection result.
Second, without the pressure from obtaining dense features, this detection method can work well with a relatively low computation complexity. Thus it can meet the real-time detection requirement.
Third, as mentioned before, a multi-frame fusion method is provided to avoid the false classification caused by noise.
Experimental results confirm the efficiency of our algorithm in different scenes.

In this paper, we assume that training data and future data have the same feature space.
However, \cite{Shao2014} mentioned that this assumption may not be guaranteed due to the limited availability of human labeled training data.
For this case, some methods based on transfer learning should be considered.
As the traffic scenarios we adopt to test our method here are different scenes captured by fixed cameras, the features of bicycles we select do not change severely.
As a result, the effect of the feature space shifting is not significant.
For further research or other different application scenarios, like handling videos captured by cameras with PAN/tilt, feature space shifting should be taken into consideration.

The rest of this paper is organized as follows. In Section \ref{sect:overview}, an overview of our method is given.
Then the bicycle detection method based on multi-feature fusion is described in detail in Section \ref{sect:fusion}.
The multi-frame fusion is explained in Section \ref{sect:framefusion}.
The framework of this bicycle detection method is provided in Section \ref{sect:framework}.
Experimental results are presented in Section \ref{sect:experiment}, where three different scenes are used to verify our algorithm.
In Section \ref{sect:conclusion}, some concluding remarks are placed.
\section{Overview of our bicycle detection algorithm}
\label{sect:overview}

As mentioned in Section \ref{sect:introduction}, there are three drawbacks in current researches--the improper features for bicycle detection in traffic videos, the insufficient use of extracted features and the high computation complexity.
To conquer these shortcomings, a method based on multi-feature fusion method and multi-frame fusion method is adopted.

First, in this method, a group of simple but effective features like geometric features and velocity features are selected.
These features are easy to compute and can describe the global characteristics of the bicycles.
Unlike dense features, they are not insensitive to the quality of videos.
Thus, they can be utilized in the scenario of low-resolution traffic videos.
Second, to get a precise result, this method fuses features and judgments in two stages.
In the first stage, this method fuses the obtained features through support vector machine(SVM) or cascade classifier method to get a preliminary detection decision in the single frame level.
In the second stage, it fuse the preliminary detection results from multiple frames by the majority rule to obtain a more reliable detection result in the image sequence level.
Third, due to the low computation complexity in feature extraction and object classification, this method can achieve a high computational efficiency, which can meet the real-time requirement.
\section{\textsc{Multi-feature fusion}}
\label{sect:fusion}

There are two key factors in the multi-feature fusion, including the selection of appropriate features and
the fusion method. Selecting a group of reliable and salient features is the foundation of the multi-feature fusion
while the fusion method determines the speed and accuracy of bicycle detection.

In this section, we provide the multi-feature fusion method.
Both feature selection and multi-feature fusion method are proposed.
First, to avoid the error caused by dense features, to decrease the impact of the quality of image and to reduce the computational complexity, geometric features and velocity features are used in this method.
Second, considering both effective detection and efficiency, two methods -- SVM and cascade classifier -- are given.

\subsection{Selection of features}
\label{sect:feature}

There are vastly different features for object detection in the existing literature.
A typical example of features is the Histograms of Oriented Gradients (HOG) feature in \cite{Dalal2006},
where excellent pedestrian detection is obtained.
The HOG feature is a kind of dense feature and they need the detailed information of objects and require objects to be large enough,
which may not be satisfied by the low-resolution traffic videos.
Moreover,
the processing of dense features is usually time-consuming and cannot be done in a real-time fashion.

The geometric features mainly refer to the salient features of objects and are less sensitive to the video quality.
Although a single geometric feature may not supply enough information for object detection,
we can fuse multiple geometric features to yield good detection results. Moreover, the processing of geometric features
is fast and can be implemented to real-time applications.
We choose the following geometric features.
Note that we bound each object with a rectangular box, which is referred to as object region.

\begin{itemize}
\item{\textbf{The number of foreground pixels in an object region.}}

\item{\textbf{The width, length and aspect ratio of an object region.}}

\item{\textbf{The foreground duty cycle of an object region.}}
The foreground duty cycle of the object region $ R_f $ stands for the ratio of the number of foreground pixels over the total number of pixels in an object region.

We can further equally divide the object region into two equal halves and calculate $ R_f $ of the upper and lower sub-regions respectively, which
are denoted as $R_{f, upper}$ and $R_{f, lower}$.
For vehicles and pedestrians, $R_{f, upper}$ and $R_{f, lower}$ are close to each other while
$R_{f, upper}$ is much smaller than $R_{f, lower}$ in bicycles.

\item{\textbf{The speed of an object.}} Suppose the object has been recognized in $n$ frames.
The speed of the object is estimated as
\begin{equation}
\label{equ:vei}v_i=\frac{\sum_{j=1}^{n-1} \Delta S_j}{n-1}
\end{equation}
where $ \Delta S_j $ is the displacement of the centers of the object's bounding rectangles between two consecutive frames.
The above average method can effectively reduce the effects of the inaccurate foreground segmentation on the speed estimation.
\end{itemize}
\subsection{The multi-feature fusion method based on linear SVM}
\label{sect:SVM}

As mentioned in the last subsection, we extract multiple geometric features. A single feature cannot
produce satisfactory bicycle detection results. So we consider to fuse these features together to
detect bicycles more reliably. One multi-feature fusion method is linear support vector machine(SVM)\cite{Zhang2000},
which is widely implemented in image processing due to its excellent classification performance.
SVM constructs the maximum-margin hyperplane to separate data sets as widely as possible.

Now we explain the procedure of our multi-feature fusion based on linear SVM.
We first extract a large number of bicycle images from the videos to construct the positive sample set.
Then we extract non-bicycle images from the videos (approximately twice of bicycle images)
to construct the negative sample set. The feature vector of each sample in the positive and negative sample sets
is denoted as
\begin{equation}
\label{equ:SVMlabel}x=(feature_1,feature_2,feature_3,...,feature_n)',
\end{equation}
where $ feature_1 $, $ feature_2 $, ..., $ feature_n $ represent the value of each feature, and $x$ is a column feature vector.
The positive samples are expected to be separated from the negative samples by the following hyperplane,
\begin{equation}
\label{equ:SVMhp}w\cdot x-b=0,
\end{equation}
where $x$ is the feature vector, $w$ is the normal vector of the hyperplane and is a row vector, and
$''\cdot''$ represents the product of two vectors (a row vector multiplies a column vector).
We train SVM with the positive and negative samples and get $w$ and $b$.
The features can be fused as follows,
\begin{equation}
\label{eq:svm}
F=w\cdot (feature_1,feature_2,...,feature_n)'-b,
\end{equation}
where $ F $ is the fusion result. With $F$, we make the following final decision,
\begin{eqnarray}
\label{eq:svm_F}
\left\{
\begin{array}{ll}
\textrm{An object is a bicycle}, & F\geq 0\\
\textrm{An object is NOT a bicycle}, & F < 0
\end{array} \right..
\end{eqnarray}
By the multi-feature fusion in eq. \ref{eq:svm} and \ref{eq:svm_F},
we can use these features more effectively and the final classification result is more desirable.
\subsection{The multi-feature fusion method based on the cascade classifier}
\label{sect:cascade}

\begin{figure}[!ht]
\centering
\includegraphics[width = \linewidth]{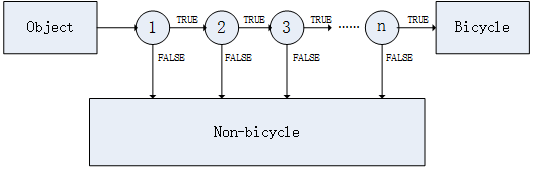}
\caption{The structure of cascade classifier}
\label{fig:cascade}
\end{figure}

Fusing features with the cascade classifier also can obtain high accuracy and low complexity.
It can be regard as a series of single feature classifier. The vague but easy accessible classifiers put in front stages, and the precise and difficult classifiers put in back stages.
Cascade classifier\cite{Viola2004} is a combination of simple classifiers with a series structure,  which is shown in Fig. \ref{fig:cascade},

In Fig. \ref{fig:cascade}, each circle represents a classifier, which uses a single feature to determine
whether an object is possibly a bicycle. When a classifier believes an object is NOT a bicycle,
it yields the {\it FALSE} decision and the detection of that object is terminated.
As mentioned before, a classifier is built upon a single feature and could make a mistake.
Therefore, the weak classifiers at all levels are combined in a series to obtain a strong overall classifier.
It is beneficial to place simple classifiers at the beginning levels of the cascade classifier because
these classifiers will exclude some objects and save the computational time of the subsequent (complicated)
classifiers. Cascade classifier has very fast computational speed and is applicable for real-time bicycle detection.

In our algorithm, we place the simple classifiers based on the shape information of objects, such as the width, length and aspect ratio,
at the beginning stages of the cascade classifier. These simple classifiers can determine that some objects are not bicycles
and exclude them. Finer detection, e.g., distinguishing a bicycle from a pedestrian, is achieved
by more complicated classifiers, like the speed of an object, at the later stages of the cascade classifier.
\section{\textsc{Multi-frame fusion}}
\label{sect:framefusion}

In Sections \ref{sect:SVM} and \ref{sect:cascade}, we fuse the features extracted from a single frame
to obtain a preliminary detection result for an object. As we know, a single frame can be disturbed by noise and/or light changing
so that the detection decision from that frame could be wrong. If we combine the detection results from
multiple frames together, the detection errors in a few frames may not be that serious, which exactly inspires
our multi-frame fusion. Generally speaking, an object $A$ is finally determined to be a bicycle if the bicycle detection decision
is made in most of the frames where $A$ is detected. Moreover, the number of frames in which $A$ is detected
provides a way to quantify the reliability of the final decision, which is referred to as confidence level.
As we show later in this section, confidence level can effectively balance the detection accuracy and the false alarm rate.
\subsection{Fusing rule}
\label{sect:rules}

As mentioned in Section \ref{sect:fusion}, a bicycle is not too different from a pedestrian, especially under
the disturbance of noise and/or light change. Although the multi-feature fusion in a single frame can
reduce such error probability, that reduction is not enough and we need the multi-frame fusion.

Our multi-frame fusion follows the majority rule. Suppose an object $A$ has been detected in $M$ frames, which may not be consecutive.
Among them, $M_b$ frames make the bicycle decision regarding $A$. Then the multi-frame fusion makes the following decision,
\begin{eqnarray}
\label{eq:framefusion}
\left\{ \begin{array}{ll}
\textrm{$A$ is a bicycle}, &M_b > M/2\\
\textrm{$A$ is NOT a bicycle}, &otherwise
\end{array} \right..
\end{eqnarray}
When $M$ is large enough, to say $M \geq 10$, the multi-frame fusion in eq. \ref{eq:framefusion} is
quite robust against noise. When $M$ is small, the reliability of the multi-frame fusion is weak.
The reliability of a decision will be quantitatively represented by confidence level, which is described in detail in Section \ref{sect:confidencelevel}.
\subsection{Life cycle and confidence level}
\label{sect:confidencelevel}
We first introduce the concept of life cycle. Life cycle is a threshold on the number of frames, and is
denoted as $N$. Suppose an object $A$ in the stored object set cannot match with any segmented object.
Then the life of $A$ is increased by $1$. When the life of $A$ is larger than a given life cycle, $N$,
$A$ is kicked out of the stored object set because it is believed to have already left the detection region.

The motivation of life cycle is to improve the robustness against disturbance. Due to noise or light change,
an object $A$ may not be correctly segmented in a frame and cannot find any match.
Suppose $A$ is kicked out of the stored object set immediately after no matching.
After several frames, $A$ may be segmented correctly, but detected as a new object,
which yields that the number of reported objects is much larger than the number of real objects,
i.e., the so-called {\it ``duplicated detection''}. Due to duplicated detection,
an object is reported as several ones, which significantly reduces the efficiency of object detection and
increases the burden on the object database. In order to resolve this issue, the kick-out of an object
is delayed by $N$ frames, i.e., an object is kicked out of the stored object list if it cannot
find any match in $N$ consecutive frames. In that case, duplicated detection can be efficiently attenuated.

In our experiments, we choose $N$ around $15$. Our frame rate is $25\,\, \textrm{frame/second}$.
$N=15$ frames take about $0.6$ seconds, during which a bicycle will not move too much and
the delayed matching still makes sense. When $N$ is too large, the movement of an object can be
large during $N$ frames and two different objects may be matched by mistake.

In the multi-frame fusion rule in eq. \ref{eq:framefusion}, an object $A$ has been detected in $M$ frames.
The larger is $M$, the more reliable is the detection result.
According to $M$, we can define the confidence level of a decision, $ COF $,  as
\begin{equation}
\label{dfn:cof}
COF=\left\{ \begin{array}{ll}
1, &M \geq N\\
\frac{M}{N}, &M < N
\end{array} \right..
\end{equation}
$COF$ measures the reliability of object detection results. When $COF$ is close to $1$,
we are more confident about the detection decision. When $COF$ is low,
we are less sure about the obtained result.

We introduce a threshold on $COF$, $T_{COF}$,
to decide whether a detected object $A$ is acceptable by
\begin{equation}
\label{equ:cof}
\left\{ \begin{array}{ll}
\textrm{$A$ is acceptable}, & COF \geq T_{COF}\\
\textrm{$A$ is NOT acceptable}, & COF < T_{COF}
\end{array} \right..
\end{equation}
$T_{COF}$ measures the reliability of acceptable object detection results and lies between $0$ and $1$.
By setting different confidence level threshold $T_{COF}$, we can select object detection results according to the needs of users.
When $T_{COF}$ is set high (close to 1), the concerned results are required to be more accurate
while some detection results with small $COF$ will be discarded.
When $T_{COF}$ is set low, more detection results are accepted and
the false alarm rate could also be high due to the included results being detected in only a few frames.
Anyway, $T_{COF}$ provides users a way to balance among the accuracy, the false alarm rate and the completeness of
detection results.
\section{The framework of the designed bicycle detection method}
\label{sect:framework}

\begin{figure}[!ht]
\centering
\includegraphics[width = \linewidth]{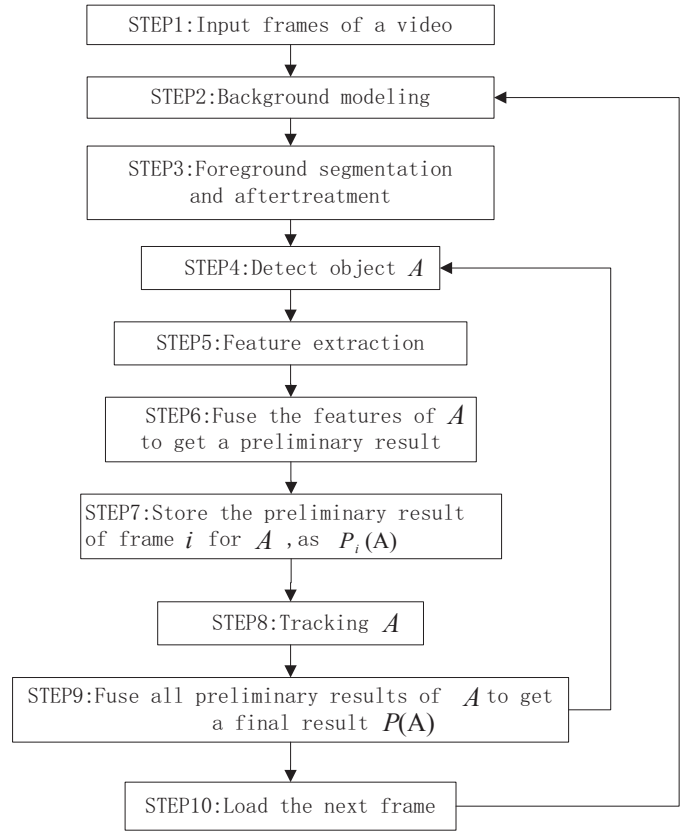}
\caption{The diagram of our bicycle detection algorithm.}
\label{fig:algorithm}
\end{figure}

The procedure of our algorithm is shown in Fig. \ref{fig:algorithm}. Now we briefly explain its main steps.

In STEP 2, the background model is obtained by the common background updating method, such as GMM.

In Step 3, the foreground is achieved through background subtraction.
There may exist some holes and noisy points in the achieved foreground due to the background noise, light change, camera shaking, etc.
So we do aftertreatment, such as erosion, dilation, basic morphology processing and target fusion, in order to get a better foreground.

In STEP 4, the processed foreground is partitioned into objects. We take an object $A$ from STEP 4 to explain the subsequent steps. Other objects follow the same procedure.

In STEP 5, we extract some geometric features of $A$ in the current $i-th$ frame, such as the aspect ratio, the duty cycle of foreground, the number of foreground pixels
and the speed. In STEP 6, support vector machine method or cascade classifier is used to produce a preliminary detection result,
denoted as $P_i(A)$ (where $i$ is the frame index).

In STEP 8, we track object $A$.
More specifically, we predict the regions, where all stored objects could lie in frame $i$ by Kalman filtering \cite{Kalman1960}.
Then we compare the actual region where $ A $ appears with the predicted regions.
If the overlapping between $A$'s region and the predicted region of one object is large enough,
we decide that $ A $ and that object match, i.e., they are the same one; otherwise, $A$ is determined as a newly emerging object.
By computing the distance between the centers of
the predicted and actual regions, we can also obtain the movement information of $A$. Note that if a saved object
cannot match with any new object for more than $N$ consecutive frame, we determine that object has left the detection area
and remove it from the object list.

In STEP 9, we fuse all preliminary detection results regarding $A$ from all previous frames by the majority rule,
which provide not only a more reliable detection result, but also the confidence level of that detection result.
The procedure between STEP 4 and STEP 9 will be repeated for other objects being segmented in the current $i-th$ frame.

In Step 10, we load the next frame and repeat the above procedure.

\section{\textsc{Experimental results}}
\label{sect:experiment}

In order to validate our algorithm, we implemented and tested it with multiple traffic scenes,
which contain pedestrians, bicycles and vehicles. Our algorithm is tested with real traffic surveillance videos,
whose image size is 352*288 pixels and
frame rate is $25\,\, \textrm{frames/second}$. The algorithm is run on an ordinary PC (Intel Core i3-2120, 3.3GHz).
\subsection{Database}
\begin{figure}[!ht]
\centering
\includegraphics[width = \linewidth]{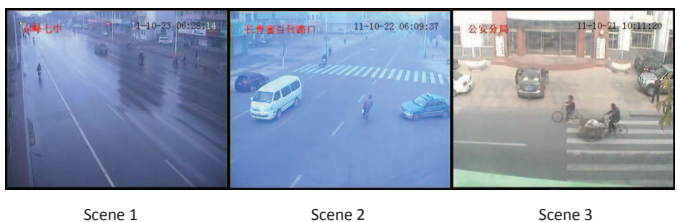}
\caption{Three different scenes}
\label{fig:scene}
\end{figure}

As there is no public dataset for bicycle detection, we generate a dataset from urban traffic videos.
We select three different scenarios, which are shown in Fig.\ref{fig:scene}.

In Fig.\ref{fig:scene}, scene 1 is a rainy day with much water on the road, scene 2 is a foggy day and
scene 3 is sunny and more favorable for detection.
\subsection{Performance indices}
\label{sect:indicators}

In the experiments, the relevant performance indices include detection rate, false alarm rate, missing rate,
duplication rate and processing time per frame, which are explained as follows.

\begin{itemize}
\item \textbf{The detection rate $ R_{det} $.}
The detection rate is the number of detected bicycles over the total number of bicycles in the video.

\item \textbf{The false alarm rate $ R_{fp} $.}
The false alarm rate is the number of objects wrongly detected as bicycles over the total number of bicycles.

\item \textbf{The duplication rate $ R_{rep} $.} It is defined as
\begin{eqnarray*}
R_{rep}=\frac{\textrm{the number of newly detected bicycles}}{\textrm{the total number of bicycles}} -1.
\end{eqnarray*}

\item \textbf{The processing time.} The processing time per frame is used to measure the
complexity of detection algorithms.

\end{itemize}
\subsection{Detection results and analysis}
\label{sect:analysis}

The experimental results for the scenes are shown in Table \ref{table:results}. We can see that
\begin{table}[htbp]
\centering
\caption{\label{table:results}Experimental results by our algorithm}
\begin{tabular}{p{0.1\linewidth}|p{0.2\linewidth}|p{0.1\linewidth}|p{0.1\linewidth}|p{0.2\linewidth}}
\hline

\hline

Scene & Method & $ R_{det} $ & $ R_{fp} $ & Processing time per frame\\

\hline

Scene1 &SVM fusion & $ 86.15\% $ & $ 16.42\% $ &  $ 30ms $\\

\hline

Scene1 &Cascade Classifier fusion & $ 87.42\% $ & $ 4.36\% $ &  $ 30ms $\\

\hline

Scene2 &SVM fusion & $ 93.88\% $ & $ 10.31\% $ &  $ 30ms $\\

\hline

Scene2 &Cascade Classifier fusion & $ 91.55\% $ & $ 6.27\% $ & $ 30ms $\\

\hline

Scene3 &SVM fusion& $ 96.96\% $ & $ 18.41\% $ &  $ 30ms $\\

\hline

Scene3 &Cascade Classifier fusion & $ 95.14\% $ & $ 8.75\% $ & $ 30ms $\\

\hline

\hline
\end{tabular}
\end{table}
\begin{enumerate}
\item In the rainy day(Scene 1), both SVM and cascade classifier methods work well with
relatively poor performance. In that scene, there exist strong shadows, which mislead the
detection of bicycles. Due to the fusion of multiple features and multiple frames, we
can still achieve reasonable results.
\item In the foggy day(Scene 2), although the moving objects are vague, we can still extract their salient features
and detect bicycles very well.
\item The missing rate of the method based on SVM is slightly lower than the one of the cascade classifier method
while the false alarm rate of the SVM method is higher than the other.
The reason is that the SVM method is built upon the simple linear combination of multiple features in eq. \ref{eq:svm},
which is easier to match objects and yields lower missing rate and higher false alarm rate.
The overall performance of the cascade classifier method is better than the one of the SVM method.
\item Both the SVM and cascade classifier methods can achieve satisfactory detection rates,
which meet the requirements of practical applications. The two methods have similar processing time per frame, which is about $30ms$.
The frame rate of ordinary videos is around 25-30 frames per second.
So these two methods can meet the real-time detection requirements of real applications.
\end{enumerate}
\subsection{The effects of the confidence level threshold}
\label{sect:repeated}

\begin{figure}[!ht]
\centering
\includegraphics[width = \linewidth]{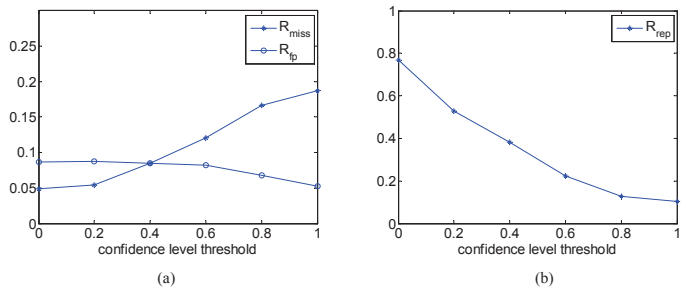}
\caption{(a)Effects of $T_{COF}$ on the missing rate and the false alarm rate; (b) Effects of $T_{COF}$ on the duplication rate.}
\label{fig:cof1}
\end{figure}

As mentioned in Section \ref{sect:confidencelevel}, we can effectively balance
the detection accuracy and the false alarm rate by selecting an appropriate confidence level threshold.
Now we try different confidence level thresholds to show
their effects on missing rate, false alarm rate and duplication rate.
Note that we take the SVM multi-feature fusion method here.
The experimental results are shown in Fig. \ref{fig:cof1} and Table \ref{table:repeated}. We can see that
\begin{table}[htbp]
\centering
\caption{\label{table:repeated}The experiment results under different confidence levels}
\begin{tabular}{p{0.15\linewidth}|p{0.08\linewidth}|p{0.08\linewidth}|p{0.08\linewidth}|p{0.08\linewidth}|p{0.08\linewidth}|p{0.08\linewidth}}
\hline
Confidence level threshold & $ 0 $ & $ 0.2 $ & $ 0.4 $ & $ 0.6 $ & $ 0.8 $ & $ 1 $\\
\hline
$ R_{det} $ & $ 95.14\% $ & $ 94.56\% $ & $ 91.54\% $ & $ 87.92\% $ & $ 83.38\% $ & $ 81.27\% $\\
\hline
$ R_{fp} $ & $ 8.70\% $ & $ 8.75\% $ & $ 8.46\% $ & $ 8.20\% $ & $ 6.76\% $ & $ 5.28\% $\\
\hline
$ R_{rep} $ & $ 76.83\% $ & $ 52.72\% $ & $ 38.28\% $ & $ 22.34\% $ & $ 12.68\% $ & $ 10.41\% $\\
\hline
\end{tabular}
\end{table}
\begin{enumerate}
\item Before considering confidence level, i.e., setting the confidence level threshold at $0$,
the duplication rate of bicycles is about $76.83\%$, which means that each bicycle has been detected
as new ones about twice in average. As the confidence level threshold increases, the duplication rate greatly decreases.
When the confidence level threshold is set at $1$, the duplication rate is $10.41\%$, which is only $15\%$ of the duplication
rate under the confidence level threshold of $0$. So we can effectively attenuate the duplication rate based on
confidence level.

\item When the confidence level threshold is set low, more detection results are accepted and the false alarm rate could also be high
due to the inclusion of detection results being obtained from a few frames.
Accordingly, the missing rate decreases as the confidence level threshold decreases.

\item In practical applications, the confidence level threshold can be adjusted
to reach a trade-off among missing rate, false alarm rate and duplication rate
according to users' preference.
When the system requires high accuracy, the confidence level threshold is set high to effectively reduce false alarms.
When the system requires low missing rate, the confidence level threshold is set low to accept more detection results,
whose reliability may be low.
\end{enumerate}
\section{Conclusion}
\label{sect:conclusion}

We present an approach based on multi-feature and multi-frame fusion to detect bicycles.
Our approach extracts the sparse geometric features (simple salient features) of objects,
and fuses these sparse geometric features by the SVM method and the cascade classifier method to improve detection performance.
The detection results from multiple frames are fused together to further reduce detection errors.
Moreover, we introduce the confidence level of detection results to achieve a desired balance among detection accuracy, false alarm rate,
missing rate and duplication rate.
As experimental results show, our approach can efficiently detect bicycles with low computational complexity,
and is therefore applicable for real-time traffic surveillance systems.





\begin{thebibliography}{00}
    \bibitem{Sheng2010}N. Sheng, H. Wang and H. Liu, ``Multi-traffic objects classification using support vector machine,'' In \emph{Proceedings of Chinese Control and Decision Conference (CCDC)}, pp. 3215-3218, May 2010.
    \bibitem{WangYubo2007}Y. Wang, Y. Wang, ``Inducement of Energy Crisis of China and the Strategy in Response to the Crisis,'' In \emph{Sino-Global Energy}, pp. 15-18, vol. 12, 2007.
    \bibitem{Ministry2011}The Ministry of Public Security of China, ``The National Road Traffic Accidents in the First Half of 2011,'' http://www.mps.gov.cn/n16/n1282/n3553/2921474.html, 2011.
    \bibitem{Dalal2006}N. Dalal and B.  Triggs, ``Histograms of oriented gradients for human detection,'' In \emph{Proceedings of the 18th Conference on Computer Vision and Pattern Recognition (CVPR 2005)}, San Diego, CA, USA, volume 1 of IEEE, pp. 886-893, 2005.
    \bibitem{Shashua2004}A. Shashua, Y. Gdalyahu and G. Hayun, ``Pedestrian Detection for Driving Assistance Systems: Single-frame Classification and System Level Performance,'' In \emph{Proceedings of IEEE Intelligent Vehicles Symposium(IV2004)}, University of Parma, pp. 1-6, 2004.
    \bibitem{Sjt2010}W. Wan, H. Huo and Y. Zhao, ``Target Detection and Recognition in Intelligent Video Surveillance,'' Shanghai Jiaotong University Press, 2010.
    \bibitem{WuYun2010}Y. Wu, Q. Kong, Z. Liu and Y. Liu, ``Pedestrian and Bicycle Detection and Tracking in Range Images,'' In \emph{Proceedings of International Conference on Optoelectronics and Image Processing(ICOIP2010)}, pp109-112, 2010.
    \bibitem{Niwat2011}N. Thepvilojanapng, K. Sugo, Y. Namiki and Y. Tobe, ``Recognizing Bicycling States with HMM based on Accelerometer and Magnetometer Data,''
In \emph{Proceedings of SICE Annual Conference}, Waseda University, pp. 831-832, 2011.
    \bibitem{Rogers1999}S. Rogers, P. Nikolaos, ``A Robust Video-based Bicycle Counting System,'' In \emph{Proceedings of ITS America Meeting (9th: New thinking in transportation)}, Washington DC, pp. 1-12, 1999.
    \bibitem{ChungChengChiu2007}C. Chiu, M. Ku and H. Chen, ``Motorcycle Detection and Tracking System with Occlusion Segementation,'' In \emph{Proceedings of the Eight International Workshop on Image Analysis for Multimedia Interactive Services(WIAMIS'07)}, pp.  32-35, 2007.
    \bibitem{Cho2010}H. Cho, P. Rybski and W. Zhang, ``Vision-based Bicyclist Detection and Tracking for Intelligent Vehicles,'' In \emph{Proceedings of IEEE Intelligent Vehicle Symposium}, University of California, San Diego, pp. 454-461, Jun. 2010.
    \bibitem{Jung2012}H. Jung, Y. Ehara, J. K. Tan, H. Kim, and S. Ishikawa, ``Applying MSC-HOG Feature to the Detection of a Human on a Bicycle,'' In \emph{Proceedings of the 12th International Conference on Control, Automation and Systems (ICCAS)}, IEEE, pp. 514-517, 2012.
    \bibitem{Fujimoto2013}Y. Fujimoto, J. Hayashi, ``A method for bicycle detection using ellipse approximation, '' In \emph{19th Korea-Japan Joint Workshop on Frontiers of Computer Vision(FCV)}, IEEE, pp. 254-257, 2013.
    \bibitem{Moro2011}A. Moro, E. Mumolo, M. Nolich, K. Umeda, ``Real-time GPU implementation of an improved cars, pedestrians and bicycles detection and classification system, ''In \emph{14th International IEEE Conference on Intelligent Transportation Systems (ITSC)}, IEEE, pp.1343-1348, 2011.
    \bibitem{Shao2014}L. Shao, F. Zhu, and X. Li. ``Transfer learning for visual categorization: A survey, ''In \emph{IEEE Transactions on Neural Networks and Learning Systems},2014.
    \bibitem{Kalman1960}R. Kalman, ``A new approach to linear filtering and prediction problems,''
In \emph{Transactions of the ASME Journal of Basic Engineering}, pp. 35-45, 1960.
    \bibitem{Zhang2000}Z. Bian, X. Zhang, ``Pattern Recognition'', pp. 284-299, Tsinghua University Press, 2000.
    \bibitem{Viola2004}P. Viola, ``Robust Real-time Face Detection'', \emph{International Journal of Computer Vision}, vol. 57, No. 2, pp. 137-152, 2004.
    \bibitem{Zhang_ccc}
    Y. Zhang, J. Yan, Q. Ling, F. Li and J. Zhu. Moving cast shadow detection based on regional growth. In Control Conference (CCC), 2013 32nd Chinese, pp. 3791-3794, IEEE, 2013.
    \bibitem{Yan_jcis}
    J. Yan, Q. Ling, Y. Zhang, F. Li and F. Zhao. An adaptive bicycle detection algorithm based on multi-Gaussian models. Journal of Computational Information Systems, 9(24), pp.10075-10083, 2013.
    \bibitem{Yan_ccc}
    J. Yan, Q. Ling, Y. Zhang, F. Li and F. Zhao. A novel occlusion-adaptive multi-object tracking method for road surveillance applications. In Control Conference (CCC), 2013 32nd Chinese, pp. 3547-3551, IEEE, 2013.
\end{thebibliography}







\end{document}